\documentclass{article}

\usepackage{arxiv}

\usepackage[utf8]{inputenc} % allow utf-8 input
\usepackage[T1]{fontenc}    % use 8-bit T1 fonts
\usepackage{hyperref}       % hyperlinks
\usepackage{url}            % simple URL typesetting
\usepackage{booktabs}       % professional-quality tables
\usepackage{amsfonts}       % blackboard math symbols
\usepackage{nicefrac}       % compact symbols for 1/2, etc.
\usepackage{microtype}      % microtypography
\usepackage{lipsum}		% Can be removed after putting your text content
\usepackage{graphicx}
\usepackage{natbib}
\usepackage{doi}
\usepackage{multirow}

\title{Self Generated Wargame AI: Double Layer Agent Task Planning Based on Large Language Model}

\author{ \href{https://orcid.org/0000-0000-0000-0000}{\includegraphics[scale=0.06]{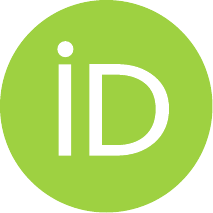}\hspace{1mm}Yuxiang Sun} \\
	School of Management and Engineering\\
	Nanjing University\\
	Nanjing, 210023 \\
	\texttt{sunyuxiang@nju.edu.cn} \\
	\And
    \href{https://orcid.org/0000-0000-0000-0000}{\includegraphics[scale=0.06]{orcid.pdf}\hspace{1mm}Junjie Zhao} \\
	School of Management and Engineering\\
	Nanjing University\\
	Nanjing, 210023 \\
	\texttt{junjiezhao@smail.nju.edu.cn} \\
    \And
	\href{https://orcid.org/0000-0000-0000-0000}{\includegraphics[scale=0.06]{orcid.pdf}\hspace{1mm}Checheng Yu} \\
	School of Management and Engineering\\
	Nanjing University\\
	Nanjing, 210023 \\
	\texttt{211870228@smail.nju.edu.cn} \\
    \And
	\href{https://orcid.org/0000-0000-0000-0000}{\includegraphics[scale=0.06]{orcid.pdf}\hspace{1mm}Wei Wang} \\
	School of Intelligence Science and Technology\\
	Nanjing University\\
	Nanjing, 210023 \\
	\texttt{221900255@smail.nju.edu.cn} \\
        \And
	\href{https://orcid.org/0000-0000-0000-0000}{\includegraphics[scale=0.06]{orcid.pdf}\hspace{1mm}Xianzhong Zhou}\thanks{This work was supported by the National Natural Science Foundation of China under Grant 61876079. (Corresponding author: Xianzhong Zhou)} \\
	School of Management and Engineering\\
	Nanjing University\\
	Nanjing, 210023 \\
	\texttt{zhouxz@nju.edu.cn} \\
}

\renewcommand{\shorttitle}{}

\hypersetup{
pdftitle={Self Generated Wargame AI: Double Layer Agent Task Planning Based on Large Language Model},
pdfsubject={cs,AI},
pdfauthor={Yuxiang Sun, Checheng Yu, Junjie Zhao, Wei Wang and Xianzhong Zhou},
pdfkeywords={First keyword, Second keyword, More},
}

\begin{document}
\maketitle

\begin{abstract}
The large language models represented by ChatGPT have a disruptive impact on the field of artificial intelligence. But it mainly focuses on natural language processing, speech recognition, machine learning and natural language understanding. This paper innovatively applies the large language model to the field of intelligent decision-making, places the large language model in the decision-making center, and constructs an agent architecture with the large language model as the core. Based on this, it further proposes a two-layer agent task planning, issues and executes decision commands through the interaction of natural language, and carries out simulation verification through the wargame simulation environment. Through the game confrontation simulation experiment, it is found that the intelligent decision-making ability of the large language model is significantly stronger than the commonly used reinforcement learning AI and rule AI, and the intelligence, understandability and generalization are all better. And through experiments, it was found that the intelligence of the large language model is closely related to prompt. This work also extends the large language model from previous human-computer interaction to the field of intelligent decision-making, which has important reference value and significance for the development of intelligent decision-making.
\end{abstract}

% keywords can be removed
\keywords{Reinforcement learning \and Large language models \and Agents \and Generative AI}

\section{Introduction}
Since ChatGPT was officially launched on November 30, 2022, it has quickly become one of the most popular intelligent Chatbot \cite{van2023chatgpt,stokel2023chatgpt}. Since its inception, ChatGPT has been applied in multiple fields such as code correction \cite{surameery2023use}, public health \cite{biswas2023role}, and global warming \cite{biswas2023potential}. In July 2023, OpenAI released the Code Interpreter plugin, further enhancing ChatGPT's data parsing capabilities and addressing the natural weaknesses of large language models in mathematics and language. These developments have provided new inspiration for improving the intelligence and generalization of AI in the field of intelligent wargame simulation, that is, using ChatGPT self generated AI to make intelligent decisions in war games.

Although the development and application of rule AI and data-driven AI \cite{cheng2021knowledge}is the starting point in the field of intelligent wargame, data-driven AI has gradually become a research hotspot in recent years, in which Reinforcement learning AI has made a series of breakthroughs. In terms of data-driven AI, Liu Man, Zhang Hongjun, and others have designed a wargame decision-making framework that balances rules and data \cite{Liu2020deci}. In terms of Reinforcement learning AI, Li Chen's team from Nanjing University of Science and Technology designed a multi-agent decision-making method under the Actor Critical framework and achieved good intelligence \cite{chen2021multi}. Xu Jiale, Zhang Haidong, and others designed a CNN based strategy learning model to improve the accuracy of wargame situation prediction \cite{Xu2022}. Tencent's AI Lab used Deep reinforcement learning to achieve game confrontation in the King's Glory game, and defeated professional players \cite{ye2020mastering, tecenttowards}. In a word, with the deepening of the combination of deep learning, Reinforcement learning and intelligent wargame, the intelligence of agents has been continuously improved \cite{mnih2015human,silver2016mastering,vinyals2019grandmaster,liu2020overview}.

Although rule AI does not require a long period of training, due to its limitations in rules, the upper limit of intelligence level is difficult to break through the upper limit of rules; While data-driven AI and Reinforcement learning AI improve their intelligence and flexibility by processing large amounts of data through Reinforcement learning algorithms, their interpretability is poor, and it is difficult to achieve model migration under scenario and capture point changes \cite{sun2022most,wurman2022outracing,schrittwieser2020mastering,silver2018general}. Therefore, improving the intelligence and generalization of AI in the field of intelligent wargame becomes the focus of further research.

Moreover, the decision-making of adversarial games is complex and continuous. In order to make decisions more intelligent and generalized, this article focuses on introducing a self generated AI wargame architecture based on a large language model. Create a decision-making mechanism that involves multiple generative agents interacting, mimicking human behavior and generating interpretable, credible, and highly generalizable game adversarial intelligent decisions.
\begin{figure*}[ht]
\begin{center}
  \includegraphics[width=\textwidth]{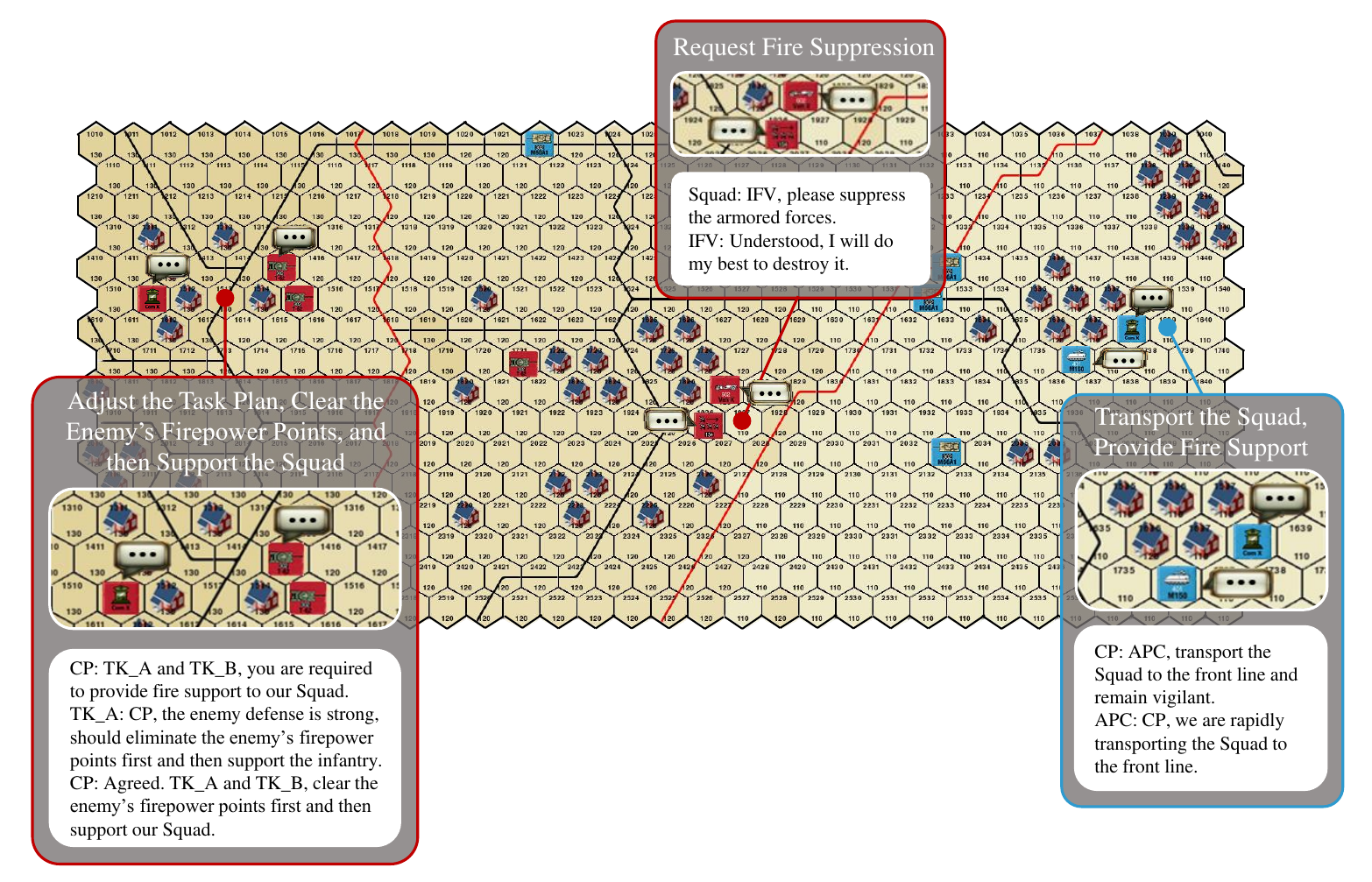}
  \end{center}
  \caption{Specific interaction of self generated wargame AI in the environment}
\end{figure*}
The core work of this article is as follows:
\begin{enumerate}
    \item The self generated AI wargame architecture is an intelligent agent architecture centered on a large language model. This architecture consists of multiple generative agents, each with its own large language model (this article uses ChatGPT as the driving tool). These intelligent agents can communicate and cooperate with each other through reflective and memory streams, and jointly make decisions. By talking to each other, they can share information, analyze the situation, and make inferences and decisions based on the conversation content.
    \item Build a two-layer agent task planning model, targeting strategic agents and tactical agents to plan tasks for the game confrontation process. Strategic agents describe specific situations observed by all current agents. 
    
    Planning refers to task allocation and execution based on all observed situational information. The tactical agent only focuses on the situation observed by a single agent chess piece and executes related tasks according to the strategic planning agent. However, tactical agents can also have their own judgments and feedback based on the prompts issued by strategic agents.
    \item Taking wargame as the experimental platform, the experiment shows that the intelligent decision-making ability of the large language model is significantly stronger than reinforcement learning AI and rule AI, and the intelligence, understandability, and generalization are all better. And through experiments. Meanwhile, research has found that providing prior knowledge from experts in the field specifically to large language models can significantly improve their intelligence.
\end{enumerate}

\section{Generative Wargame AI Architecture}

In the war chess environment, we have realized the confrontation between six red chessmen and five blue chessmen as Figure 1 shows. The red and blue chessmen in different clusters have different semantic interaction information, which is generated through ChatGPT.

To implement the decision-making mechanism mentioned above, we have developed an agent architecture consisting of three main components: a memory stream for storing and allocating buffers and generating batches; a reflection stream for using batches as prompts for the large language model to understand its role in the decision-making process; and a task planning stream for synthesizing higher-level reasoning from batches to enable the agent to integrate situational information and make better pre-battle plans. The agent architecture is designed to store, synthesize, and apply past battlefield experience to enable the large language model to generate trustworthy decisions.
\begin{figure*}
\begin{center}
    \includegraphics[width=\textwidth]{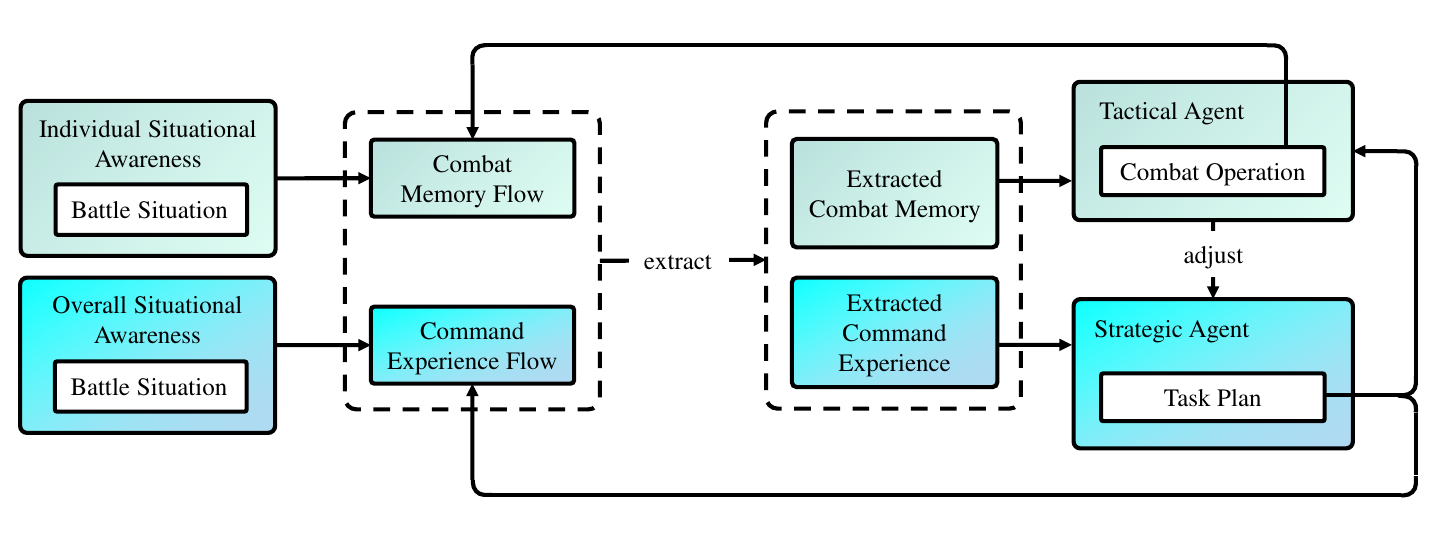}
\end{center}
  
  \caption{The relationship between strategic and tactics agent}
\end{figure*}
\section{Generative Wargame AI Model}

\subsection{Wargame Agent Interaction}
In the architecture described above, the wargame agents obtain situational information and interact with each other in natural language to maintain collaboration. Each agent describe their action in natural language, such as “red agent 1 is preparing to capture the control point and moving along the road”, “blue agent 2 is preparing to aim at the enemy target 1”. Then the sentence is translated into specific actions and directly affect the wargame environment. In the meanwhile, all the actions and movements will be displayed as a series of number symbols which appear above each avatar to provide abstract representation of actions. To achieve this, the architecture utilizes a language model to translate language into actions, while representing a concise symbol above each wargame to represent ChatGPT's suggestions for actions that this agent should take. For example, “red agent 1 is preparing to capture the control point” is displayed as “!” appearing above the wargame, while “red agent 1 is preparing to aim at the enemy” is displayed as ``→".

In this environment, agents communicate with each other in natural language which is fully understandable by humans. They obtain the situational information of other operators and environment from the semantics of sentences. Here is a sample of an agent communicating with another.
\begin{figure*}[ht]
\begin{center}
    \includegraphics[width=\textwidth]{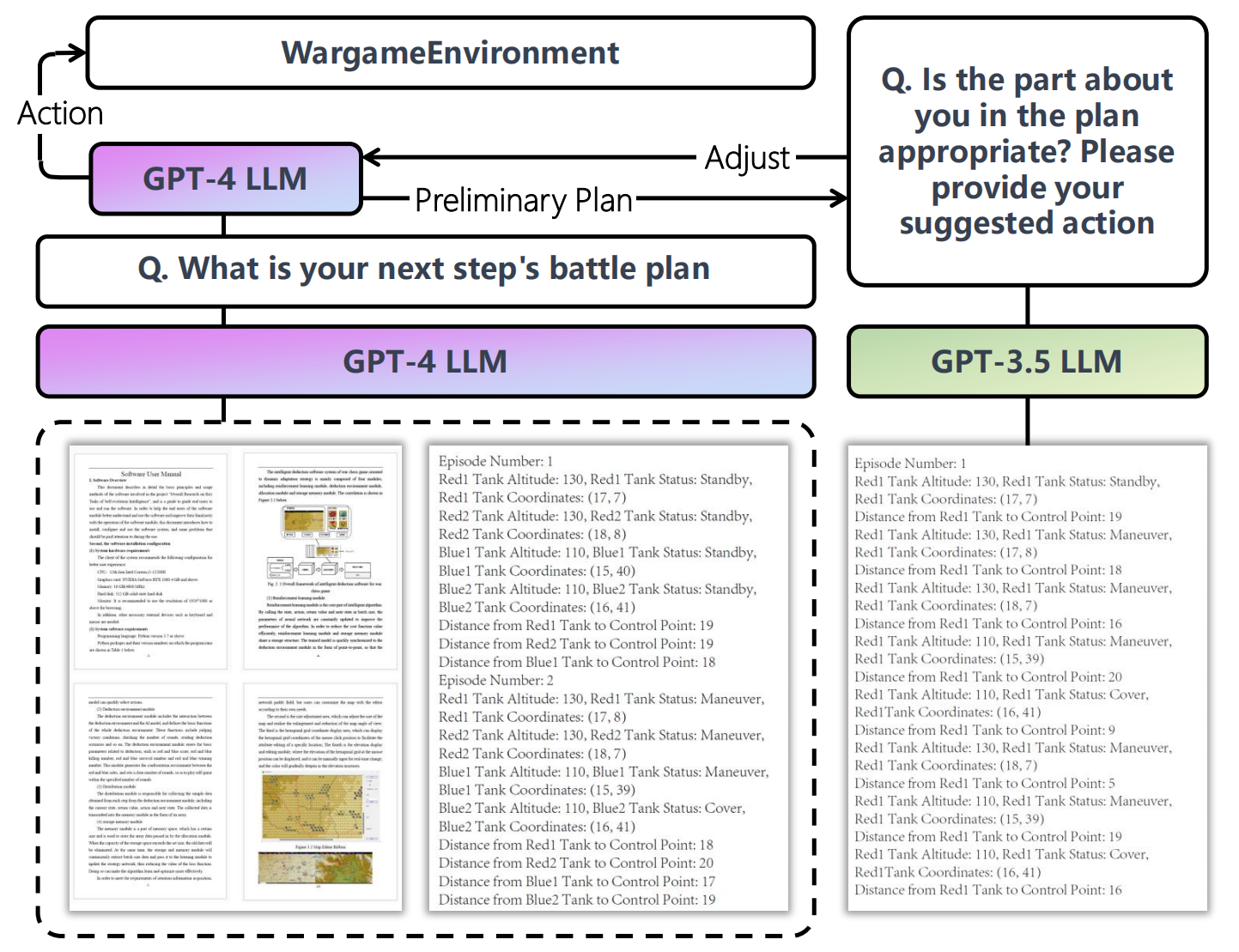}
\end{center}
  \caption{Double Layer Agent Task Planning Decision Framework Based on Large Language Model}
\end{figure*}
\subsection{Composing model}

The generative wargame AI aims to provide a novel decision-making framework for intelligent decision-making in wargame environment. Compared to the traditional rule-based AI, data-driven AI , or the reinforcement learning AI, our architecture utilizes ChatGPT for intelligent decision-making and interaction with the wargame environment. Generative wargame AI takes the current environment and past experiences as inputs, and produces output in the form of generated actions.

Generated actions can be described as these steps: the architecture provides the large language model with well-trained prompts, the language model chooses the numbers which encompass all the actions to take according to prompts, the agents take actions with the corresponding number. The innovation of the architecture lies in the combination of the large language model with retrieval of relevant information and the utilization of prompts to adjust the output of language model. 

Building upon the foundation of the architecture, we constructed a dual-layer agent system including the strategic agent and tactical agent. Strategic agent obtains all the information regarding the states of both its own sites and the observed opponents as input, then it combines this with the overall environment and input as prompts to generate a macro-level tactics intelligent task planning flow. And strategic agent assigns tasks to the tactics agent in the form of prompts, and tactics agent, based on its own states, provides modification suggestions and reasons for modifying. Then strategic agent keeps replanning according to these suggestions until all the tactics agents are not going to provide furthermore suggestions. 

Certainly, the strategic and tactics agents still face a lot of challenges even with the use of the state-of-the-art large language model like GPT-4 LLM(Large Language Model). Since extensive generation of events and memories generated by the two agents, the most critical challenge in this architecture lies in generating the most relevant memory fragments while retrieving and synthesizing relevant data from the memory stream. Therefore, this article attempts to reduce computational power and memory requirements, and uses GPT-4 LLM for strategic agent decision-making, facilitating overall strategic input and expert knowledge document input. For tactical agents, this article uses GPT-3.5 LLM for decision-making, as tactical agents can interact and provide feedback on results in turn, reducing computational power and memory requirements without affecting intelligence

\paragraph{Memory Stream}
As the central component of the architecture, the memory stream directly influences the efficiency and accuracy of decision-making. The entire memory stream is a list of memory objects, with each object consisting of a natural language description, a creation timestamp and a recent access timestamp. The fundamental element in memory list is observation, which contains all the situational information observed by agents. Due to the presence of the fog of the war, the battlefield environment doesn’t allow for complete knowledge and awareness. The common information observed by an agent in a particular state is subject to certain limitations, which includes individual actions, actions taken by our own side’s agent and the actions taken by opponent agents within our visible range.\\
Examples:\\
\emph{Observation} $1$: agent observes its own side’s agent approaching the control point and trying to control it.\\
\emph{Observation} $2$: agent observes its opposed agent approaching urban residential area and trying to shoot.

We construct a retrieval function within the entire memory stream architecture and utilize it to extract observations from the historical experiences of the agents, providing a foundation for generating reasonable prompts and enable the language model to produce rational decision. The retrieval function can be selective, with the form of prioritizing the extraction of recently observations, important nodes set before and relevant memory to produce effective outcome.

\emph{Recency} assigns a higher score to the observation added recently, in which case the agent prioritizes the memory information generated by the recent several steps. To account for the influence of time factors, we implement a time decay coefficient to calculate the score.

\emph{Importance} categorizes the data within the memory stream into regular memories and core memories, and allocating higher scores to the core memories generated by agents. For example, a red agent moving towards the left and approaching the road can be categorized as a regular memory while a red agent approaching the control point and eliminating a blue agent can be classified as a core memory. In this architecture, we ask the language model to directly output the importance integer scores in a range from $1$ to $10$, in which case $1$ means the purely common memory like moving on the road while $10$ means the most important core memory like seizing the control point or shooting successfully. The specific implementation process can be described as follows: retrieving the corresponding memory from the memory stream to form a prompt, allowing the agent to generate importance scores accordingly and storing them back to the memory stream.\\
Example:\\
Memory: the red agent one is seizing the control point.\\
\emph{Importance} score: $8$
\begin{figure*}[ht]
\begin{center}
    \includegraphics[width=\columnwidth]{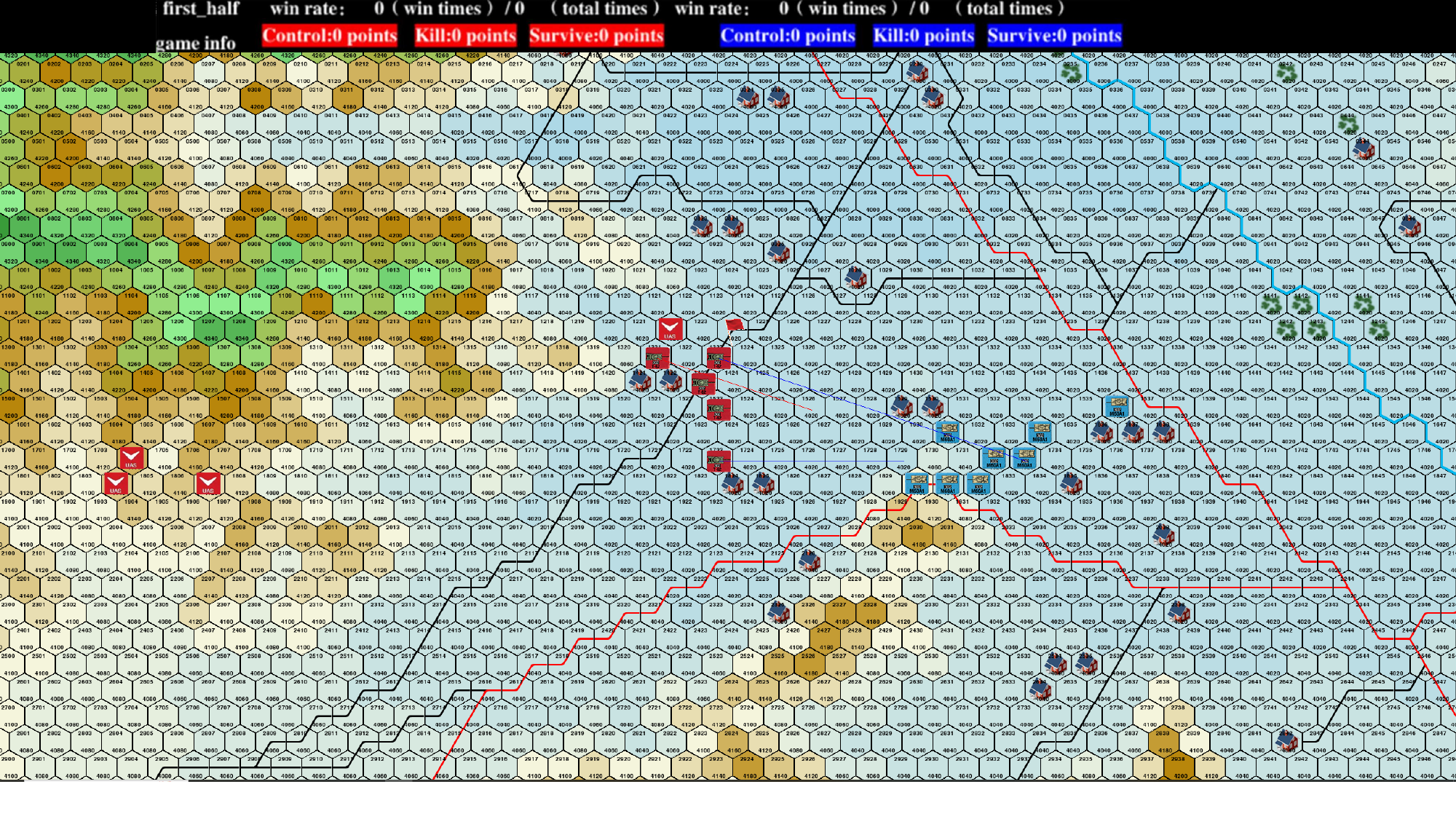}
\end{center}
  
  \caption{Experimental Simulation Environment Display}
\end{figure*}
\emph{Relevance} assigns a higher score to those object that are relevant to the current situation because of the presence of relevance between different memory objects. For example, a red agent arrives at the road and approaching the control point at a higher speed, this memory has a strong correlation with the red agent’s seizing the control point. In this paper we ask ChatGPT to generate the relevance score in a scale from 1 to 10 to describe the degree of relevance between memory objects.

As mentioned above, the three components of our architecture have been quantified into scores for the agents. To calculate the final scores of agents we normalize \emph{recency}, \emph{importance} and \emph{relevance} scores to a range of $[0,1]$ by scaling method. The final scores can be calculated using the following formula: 
$score_{final} = \alpha_{recency} * score_{recency} + \alpha_{importance} * score_{importance} + \alpha_{relevance} * score_{relevance}$

We use this score to comprehensively determine the prompts that should be extracted, and instruct the agent to generate corresponding and reasonable action-plannings based on these prompts.

\paragraph{Reflection Stream}
However, the observational performance of memory flow has limitations in the decision-making process in practical wargame environment. Reasoning based on raw observation is not efficient enough to allow a large language model to generate high-level decision results. It is necessary to infer and generate advanced reasoning semantics through the observation of information and the planning of actions. This paper defines this reasoning process as a higher-level memory flow, referred to as \emph{reflection}. It is essentially a higher-level and more abstract thinking process. The reflection flow is generated together with the memory flow, but the generation of reflection is distinguished by the retrieval function in the preceding memory flow. When the score in the retrieval function exceeds a certain threshold, reflection is triggered. This reflective process involves a higher-level abstraction and understanding of previously observed information. It is essentially a combination of observed semantics and planned semantics generated through prompts, and it is generated periodically, providing the wargame agent with reasoning semantics.

The first step of reflection is to raise questions and clarify the reflective process based on the previous experience flow of wargame agent. For example, the blue agent is approaching the road and accelerating towards the control point. The planning suggests that the red agent should reach hexagon $1403$ and shoot the blue agent at that point. From this, reflective semantics are generated: The blue agent poses a significant threat and may create a disadvantageous situation for the red agent in this confrontation.

The reflection process allows the agent to reflect not only on their current observation but on other reflection. As a result, memories generated by the agent can be divided into different levels under the reflection mechanism, allowing for more accurate decision-making at an abstract level.

\paragraph{Task Planning Stream}
Strategic agent, based on the current situation observed by all the agents of our side, describes it as a prompt following a specific format: $<\emph{Summary}, \emph{Observations}, \emph{Planning}>$. The Summary aims to convert the current situation from visual to semantic information \cite{sun2022intelligent}.

\emph{Observations} describe specific circumstances observed by all agents, further enrich semantic information based on the summary. Planning involves task allocation and execution based on the observed situation. \\
Example:\\
\emph{Summary}: our $10$ agents are moving towards the control point and have identified $3$ blue agents.
\emph{Observations}: blue agent $1$ is nearing the control point.\\
\emph{Planning}: red agent $1-3$ will prioritize engaging blue agent $1$, while agents $4-10$ will quickly move towards the control point.

\subsection{ChatGPT + Wargaming Business Process}

The whole core process is to transform the situation image information in the wargame simulation into semantic information, which includes description information and situation information, and this information is sent to the wargame agent in the form of prompt, and then the agent feedback the corresponding planning semantics, which is goal oriented. The planning semantics are then transformed into action sequences (such as $1,2,3,4,.. 10$, where numbers represent specific actions. Alternatively, they are transformed into corresponding actions such as attack, defense, evasion, acceleration, shooting, left movement, etc.), which affect the environment and generate new environments. These actions are then recycled back to the starting situation image and converted into semantics.

On this basis, in order to reduce computational power and memory requirements, and improve operational efficiency, this article allows strategic agents to use GPT-4 LLM and tactical agents to use GPT-3.5 LLM. Compared to using GPT-4 or GPT-3.5 LLM entirely, this can comprehensively improve the intelligence of intelligent decision-making without requiring too much computing power and memory space. Firstly, input expert prior knowledge documents into the strategic agent for learning through GPT-4 LLM, and then provide appropriate prompt inputs to enable the strategic agent to make decisions through GPT-4 LLM and convert them into action outputs that affect the wargame environment. The strategic agent then sends corresponding instructions to each tactical agent for execution. The tactical agent provides feedback on whether the task is suitable for the current agent through GPT-3.5 LLM combined with appropriate prompts, and provides the recommended execution results to the strategic agent for adjustment.

\emph{Strategic agent}: Based on the task planning flow, the strategic agent synthesizes states 1 to 10 and provides a task planning sequence, which is the action that each wargame should take in the step allocation;

\emph{Tactical agent}: The tactical agent receives task planning and provides modification suggestions and reasons for the assigned tasks based on its own state;

The strategic agent plans again based on the modification suggestions until all tactical agents no longer provide modification suggestions.

\section{Verification of Simulation Experimental Environment}

\subsection{Experimental Environment Display}
This paper verifies the above established large language model through simulation experiments. The simulation platform is a wargame simulation platform, which can conduct game confrontation between red and blue sides. Both red and blue sides can use intelligent algorithms to make decisions and execute actions \cite{sun2020research}. The basic adversarial rule is that the red and blue sides compete for the middle control point (red flag), and the party who first reaches the control point wins. Or if one party is completely destroyed by the other party, the other party wins.

\subsection{Advantages of large language model over reinforcement learning intelligent decision-making}
In the previous experiments, we mainly made decisions through rule AI and reinforcement learning AI. For the first time, this work used the large language model to make decisions for agents, and it was verified on this platform. Interestingly, this work found that there is a large difference between large language models and Reinforcement learning. First, large language models or trained large language models can make decisions without waiting for the convergence of training, and can directly obtain high intelligence. Reinforcement learning algorithms often need a lot of training to gradually adapt to a new task. At the same time, compared with the reinforcement learning algorithm, the decision making using the large language model can directly achieve excellent intelligence in multiple different tasks, and does not need to re-train for different tasks, which is of high value for practical applications.

This article proposes two algorithms, GWA algorithm and GWAE algorithm. The GWA algorithm adopts the composition model proposed in this article and utilizes ChatGPT for decision-making in large language models. GWAE inputs expert experience on the basis of GWA. This paper inputs expert experience of Military simulation in the form of a document. See the appendix for the document.

\begin{table}[ht]
    \centering
    \begin{tabular}{c|c|c|c}
    \hline
    \multirow{2}{*}{\textbf{Method}}& \multicolumn{3}{c}{\textbf{Missions}} \\
    \cline{2-4}
         & Kill & Goal & Survive\\
    \hline
    GWAE & 298±11 & 10504±64 & 4238±28\\
    GWA	& 332±9	& 9106±99 & 5102±33\\
    RNM-PPO	& 745±9 & 9102±141 & 4985±44\\
    PPO	& 850±19 & 7804±44 & 5068±38\\
    PK-DQN & 792±14	& 7732±60 & 5026±53\\
    DQN	& 1285±7 & 6948±161	& 5154±57\\
    \hline
    \end{tabular}
    \caption{Scores of different algorithms for three tasks: kill, get goal, and survive.}
    \label{tab:Table 1}
\end{table}
\begin{figure}[ht]
    \includegraphics[width=\columnwidth]{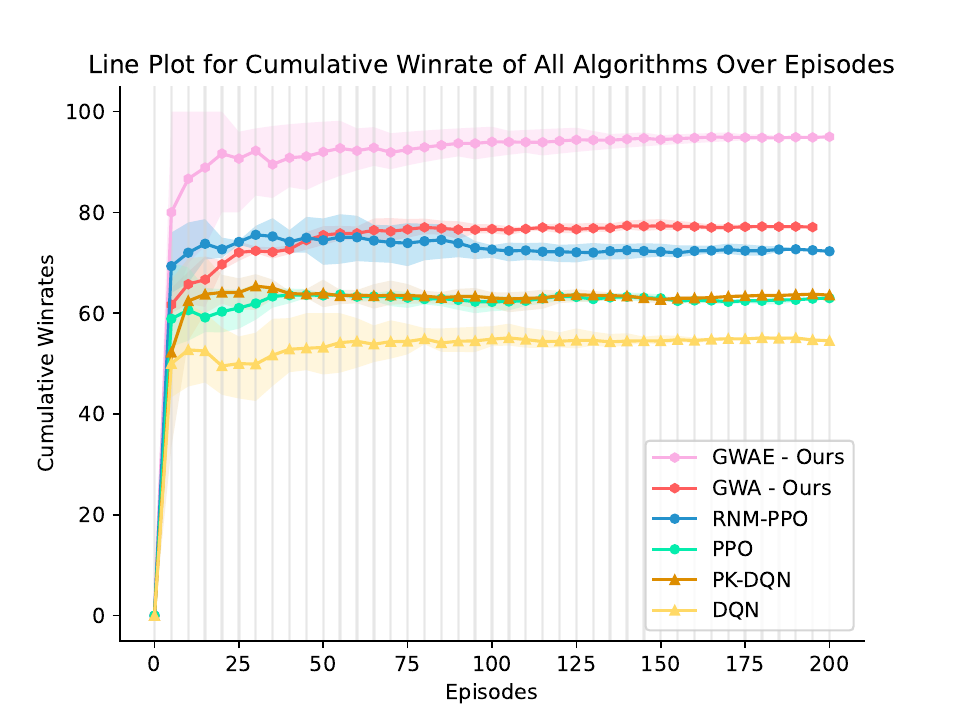}
    \caption{Line Plot for Winrate of All Algorithms Over Episodes. The winning rate of GWA algorithm is generally higher than that of Reinforcement learning algorithm, including RNM-PPO, PPO, PK-DQN, DQN. After inputting expert prior knowledge documents for GWA algorithm, the intelligence of GWAE algorithm is significantly improved on the basis of the original GWA.}
\end{figure}

\begin{figure}[t]
    \includegraphics[width=\columnwidth]{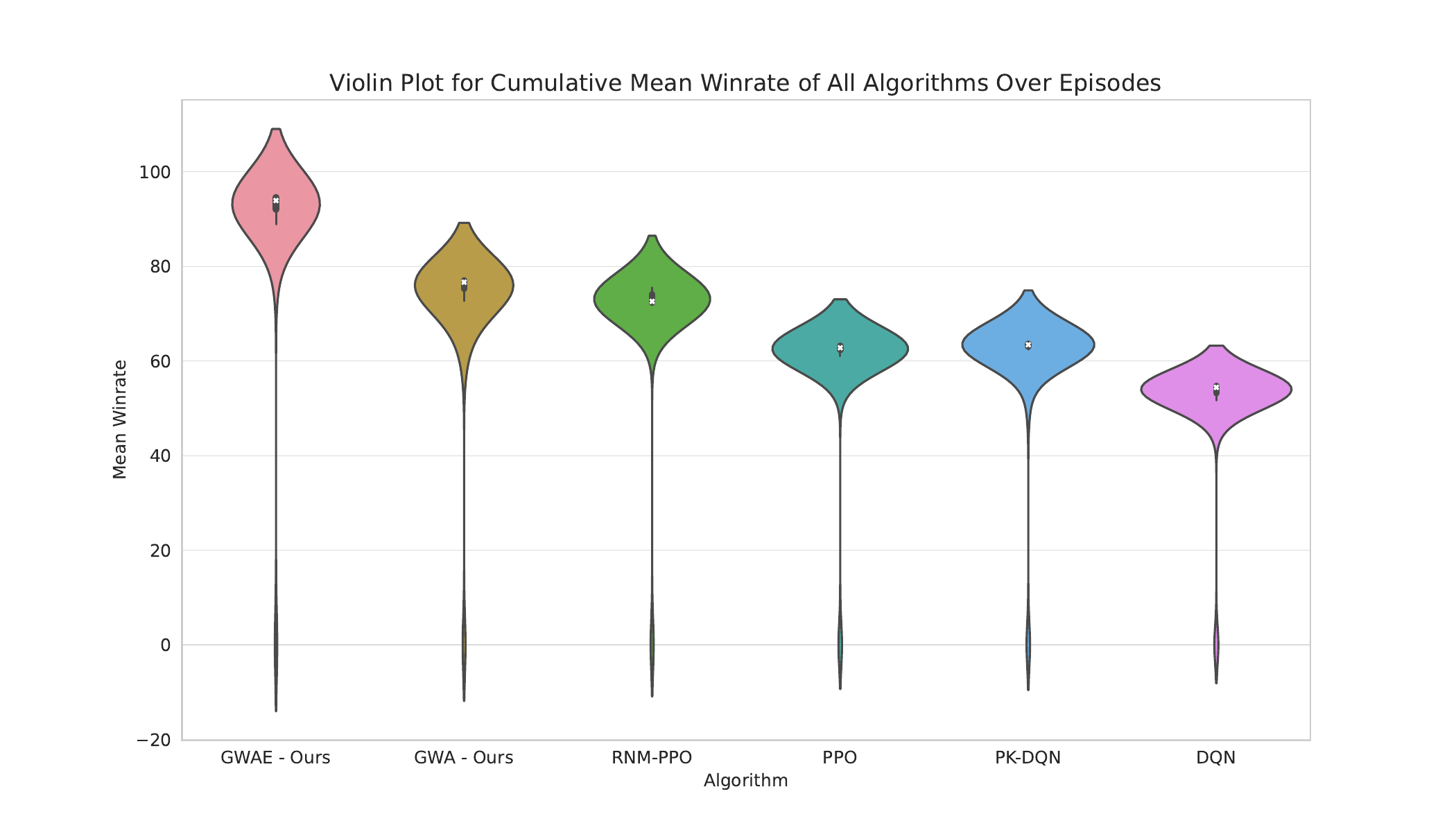}
    \caption{Violin Plot for Mean Winrate of All Algorithms Over Episodes. The GWAE algorithm and GWA algorithm have a higher winning rate and are more stable.}
\end{figure}

The experiment compares the GWAE algorithm and GWA algorithm proposed in this paper, and compares the winning rate of the algorithm proposed in this paper with the RNM-PPO \cite{xue2023multi}, PPO, PK-DQN \cite{sun2020research} and DQN algorithms. Through 
\begin{figure}[thbp!]
    \centering
    \begin{minipage}[t]{0.49\linewidth}
        \centering
        \includegraphics[width=0.9\linewidth]{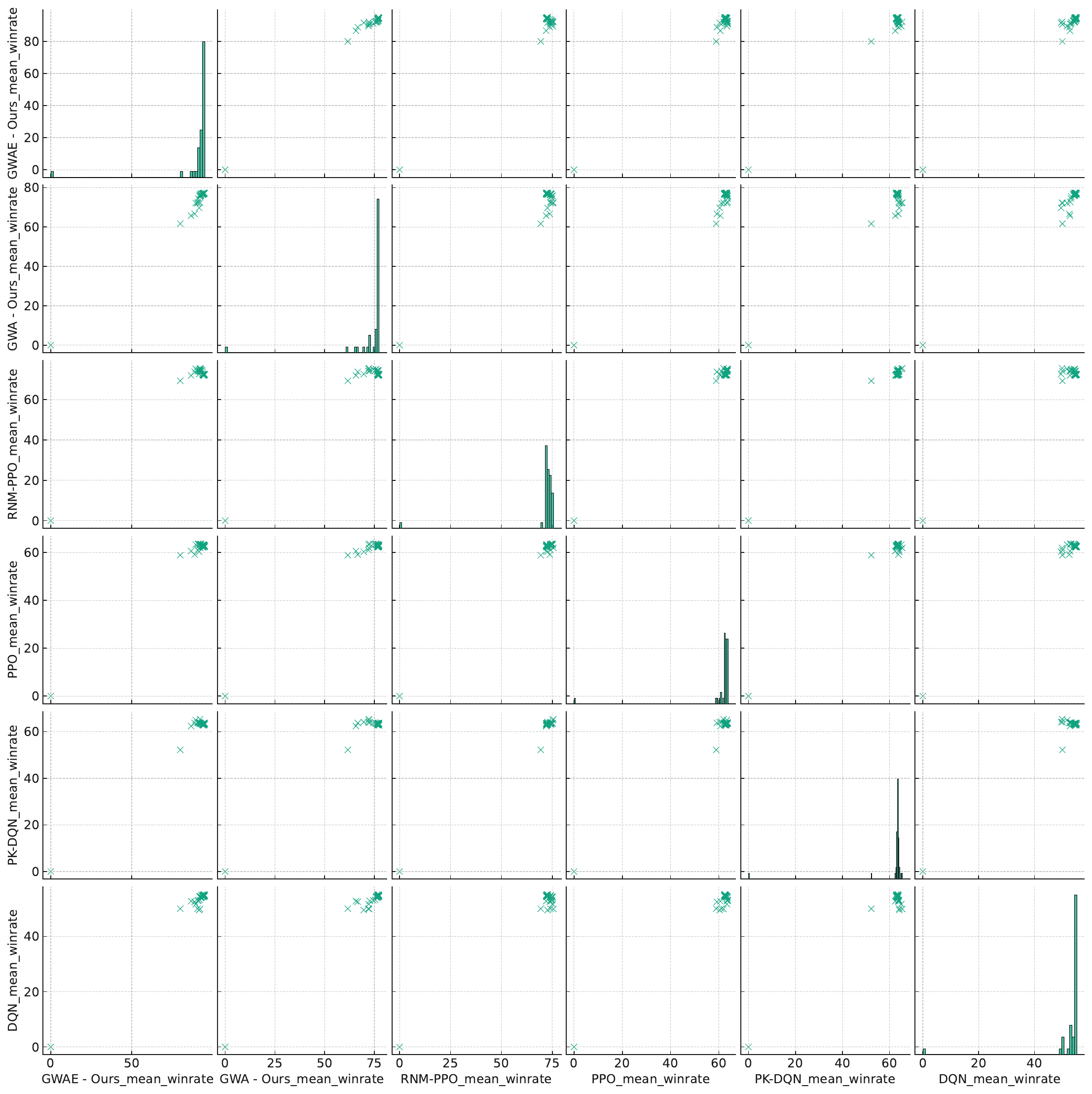}
        \caption{Scatter plot of the average winning rate of all algorithms. The results of comparing the overall winning rates of GWAE, GWA, RNM-PPO, PPO, PK-DQN, and DQN algorithms}
    \end{minipage}
    \begin{minipage}[t]{0.49\linewidth}
        \centering
        \includegraphics[width=0.9\linewidth]{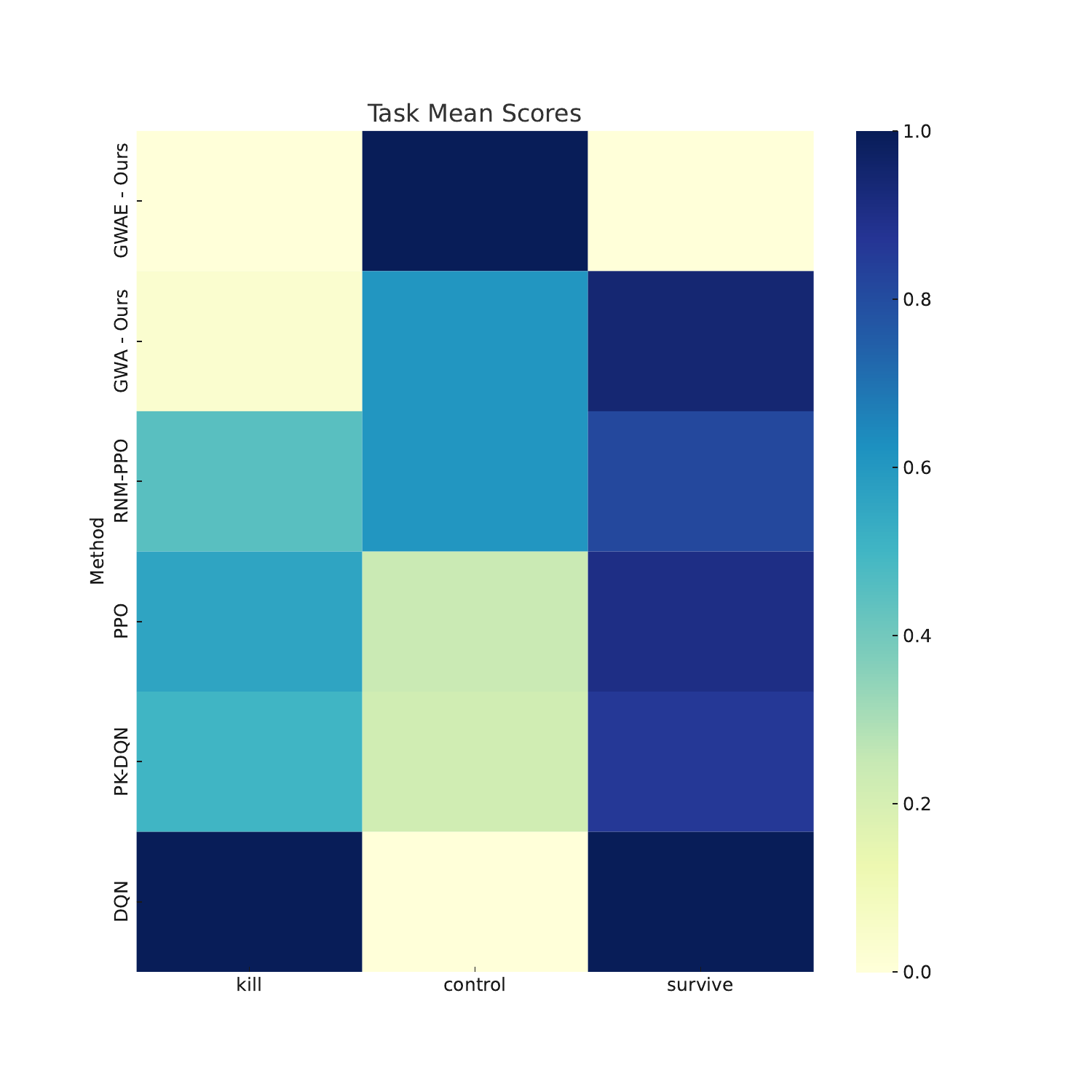}
        \caption{Task Mean Scores Heatmap.The performance of GWAE, GWA, RNM-PPO, PPO, PK-DQN, and DQN algorithms in three typical tasks: kill, control, and survive. The darker the color, the better the algorithm performs in this task.}
    \end{minipage}

 \end{figure}
 Figure 6, Figure 7 and Figure 8, it can be found that the intelligence of reinforcement learning algorithms DQN, PK-DQN, PPO, and RNM-PPO is enhanced in turn. However, the GWA algorithm that uses the large language model to make decisions is better than the 
reinforcement learning algorithm. Only the RNM-PPO algorithm is closer to GWA. If expert experience documents are fed to GWA, the GWAE algorithm's victory rate will be significantly improved. Overall, the overall winning rate of the algorithm proposed in this paper is significantly better than the previous classical reinforcement learning algorithm. Also on the premise of fixed blue intelligence and fixed reasoning scenarios, the overall effect of intelligent decision-making using the large language model is also relatively stable, the overall victory rate fluctuates relatively little. For different typical task scenarios, GWA and GWAE exhibits outstanding performance, especially in the task of scoring, the performance of GWA and GWAE is significantly better than that of classical reinforcement learning algorithms. These two algorithms have demonstrated the potential value of large language models in the field of intelligent decision-making by making appropriate decisions for different task planning scenarios without undergoing extensive training.

% \begin{figure}[ht]
%     \includegraphics[width=0.5\columnwidth]{images/Figure 7.pdf}
%     \caption{Scatter plot of the average winning rate of all algorithms. The results of comparing the overall winning rates of GWAE, GWA, RNM-PPO, PPO, PK-DQN, and DQN algorithms}
% \end{figure}

% \begin{figure}[ht]
%     \includegraphics[width=\columnwidth]{images/Figure 8.pdf}
%     \caption{Task Mean Scores Heatmap.The performance of GWAE, GWA, RNM-PPO, PPO, PK-DQN, and DQN algorithms in three typical tasks: kill, control, and survive. The darker the color, the better the algorithm performs in this task.}
% \end{figure}

\section{Conclusion}
This work innovatively applies the large language model to intelligent decision-making, and verifies the feasibility of the large language model for decision-making in the wargame platform. Compared with the intelligent decision-making of Reinforcement learning, this paper finds that the large language model has obvious advantages. Firstly, the large language model for decision-making has strong adaptability in practical game confrontations due to sufficient training in advance. There is no need to wait to restart training, and it has strong intelligence and generalization for different tasks. Secondly, the intelligence shown by the large language model is obviously stronger than the general Reinforcement learning algorithm, which proves the great potential of the large language model in decision-making. Finally, this article also found through experiments that there is a significant correlation between the intelligence of large language models and prompt. If there is a more suitable prompt, its displayed intelligence is significantly improved. Of course, the work of this article is still an initial exploration of the large language model, and there is still much work to be innovated in the future, such as the attempt of the large language model in different scenarios, and the use of the large language model in more complex game adversarial environments to further enhance the intelligence of the adversarial blue, in order to test the intelligence level of the large language model. This work also extends the large language model from previous human-computer interaction to the field of intelligent decision-making, which has important reference value and significance for the development of intelligent decision-making.

\bibliographystyle{unsrtnat}
\bibliography{sgw}

\appendix
\section{Large Language Model Prompt}
\begin{figure}[ht]
    \includegraphics[width=\columnwidth]{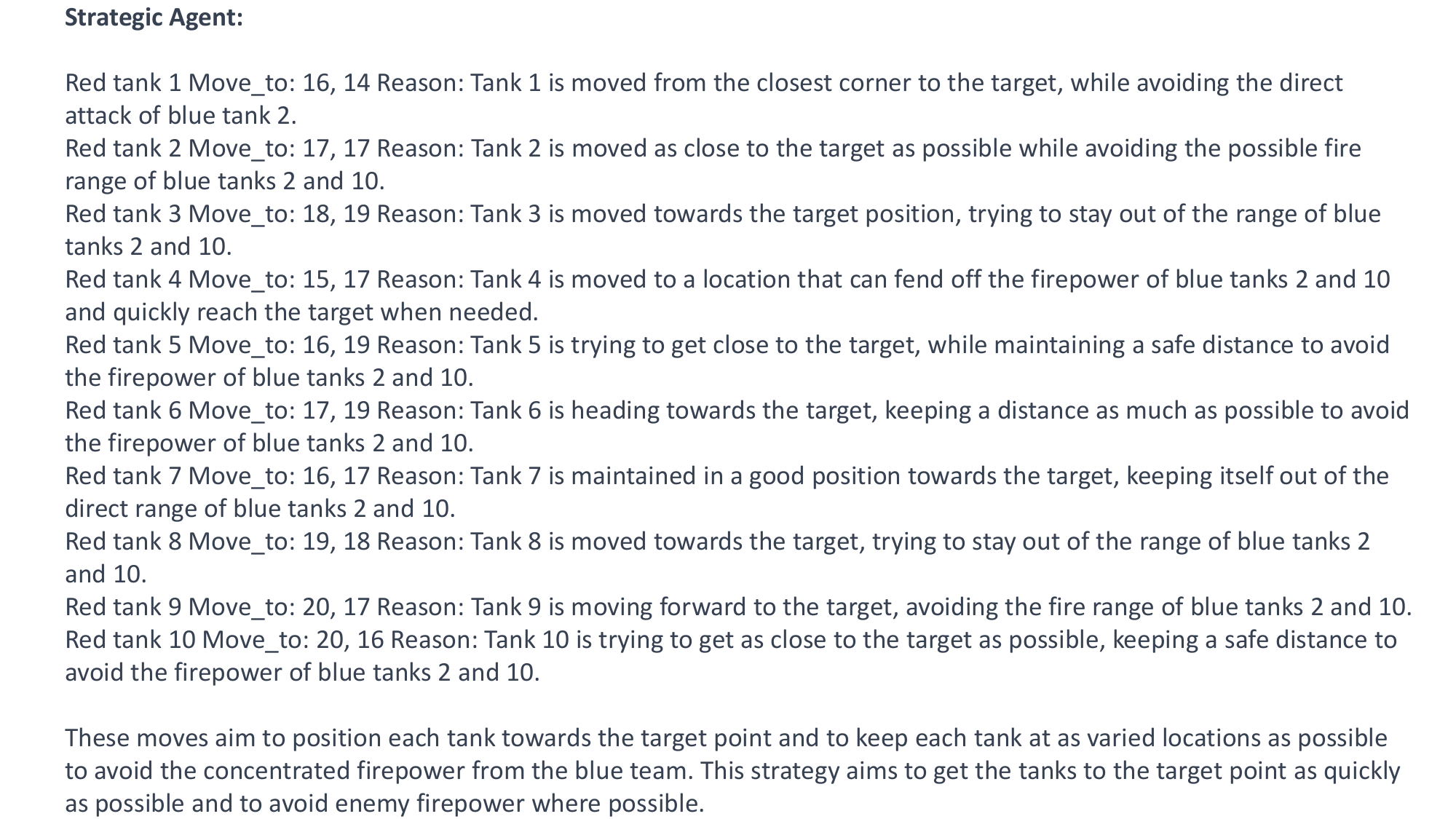}
    \caption{Strategic Agent Prompt Feedback Display}
\end{figure}

In this experiment, it was found that the intelligence of the large language model is closely related to prompt. Two different prompts were used for guidance in this experience, and the results returned after prompt input are shown in the figure. The figure 9 shows the response results of the strategic agent feedback. The figure 10 shows the response results of tactical agent feedback.

\section{Prior Knowledge Document}

This article feeds professional domain knowledge documents into the proposed GWA algorithm, forming the GWAE algorithm. Document link: https://github.com/sunyuxiang926/wargame/blob/b0118cd0\\1fa46f2790a89895336df81519d91707/files/theses/explanat\\ion\_of\_domain\_expert\_knowledge.pdf

\begin{figure}[ht]
    \includegraphics[width=\columnwidth]{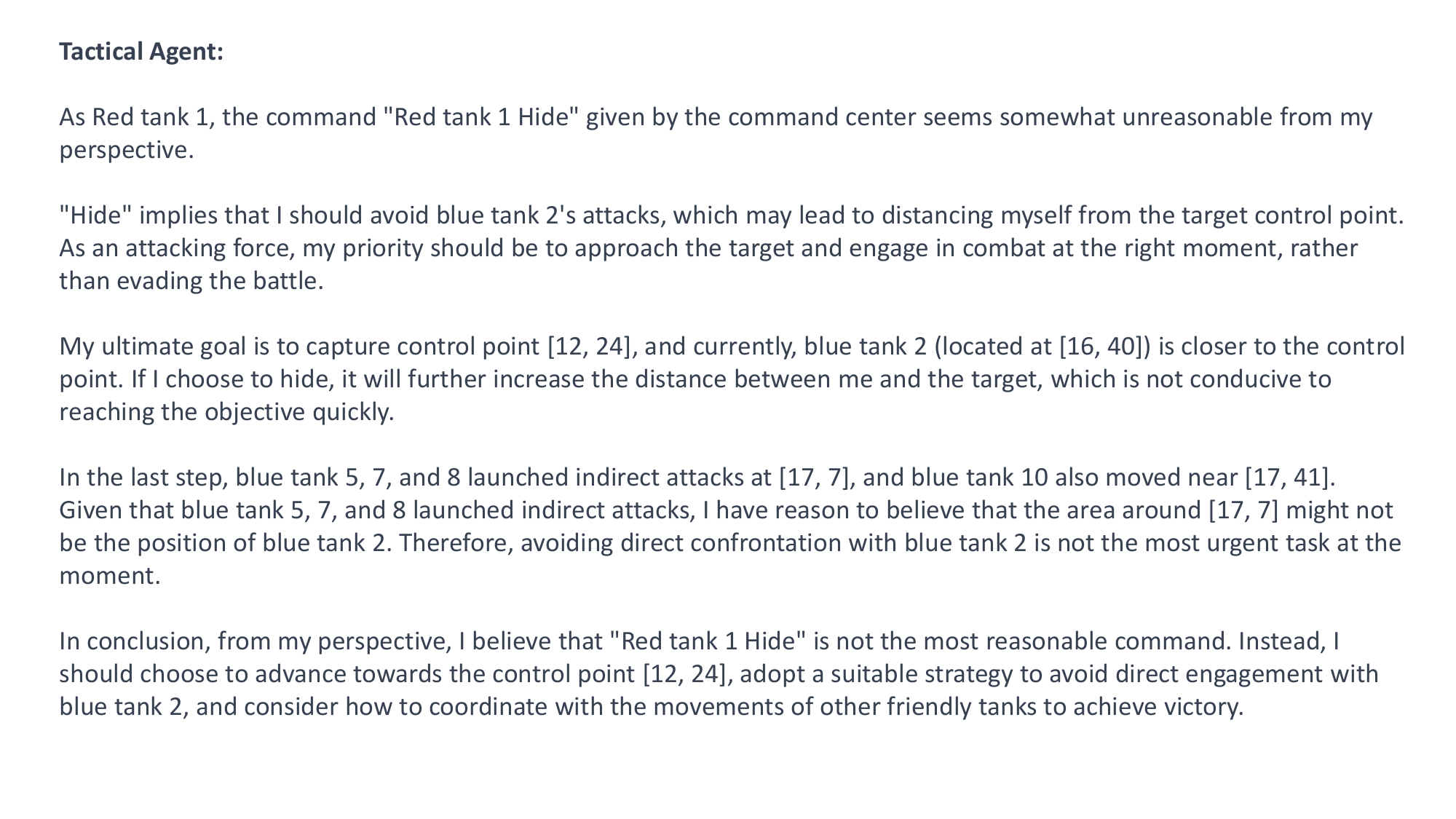}
    \caption{Tactical Agent Prompt Feedback Display}
\end{figure}
\end{document}